\pdfoutput=1
\documentclass[11pt]{article}

\usepackage[preprint]{acl}
\usepackage{bbding}
\usepackage{color}
\usepackage{times}
\usepackage{latexsym}
\usepackage{amsmath}
\usepackage{amssymb}
\usepackage{adjustbox}
\usepackage[T1]{fontenc}
\usepackage[utf8]{inputenc}
\usepackage{booktabs}
\usepackage{microtype}

\usepackage{inconsolata}

\usepackage{graphicx}

\title{ReMamba: Equip Mamba with Effective Long-Sequence Modeling}

\author{
 \textbf{Danlong Yuan\textsuperscript{1,2}\thanks{Work done during internship at Meituan}},
 \textbf{Jiahao Liu\textsuperscript{5}},
 \textbf{Bei Li\textsuperscript{5}},
 \textbf{Huishuai Zhang\textsuperscript{1,3}\thanks{Corresponding author.}},
\\
 \textbf{Jingang Wang\textsuperscript{5}},
 \textbf{ Xunliang Cai\textsuperscript{5}}
 \textbf{ Dongyan Zhao\textsuperscript{1,2,3,4}\footnotemark[2]}
\\
 \textsuperscript{1}Wangxuan Institute of Computer Technology, Peking University,
 \\
 \textsuperscript{2}Center for Data Science, AAIS, Peking University,
\\
 \textsuperscript{3}National Key Laboratory of General Artificial Intelligence,
 \textsuperscript{4}BIGAI, Beijing, China;,
 \\
 \textsuperscript{5}Meituan
\\
 \small{
   \textbf{Correspondence:} {danlongyuan@stu.pku.edu.cn},
   \{zhanghuishuai,zhaodongyan\}@pku.edu.cn,  
 }
 \\
  \small{
 \{liujiahao12,libei17,wangjingang02,caixunliang\}@meituan.com
 }
 \\
  \small{
  \textbf{Code:} \href{https://github.com/lblankl/ReMamba}{https://github.com/lblankl/ReMamba}
  }
}

\begin{document}
\maketitle
\begin{abstract}
While the Mamba architecture demonstrates superior inference efficiency and competitive performance on short-context natural language processing (NLP) tasks, empirical evidence suggests its capacity to comprehend long contexts is limited compared to transformer-based models. In this study, we investigate the long-context efficiency issues of the Mamba models and propose ReMamba, which enhances Mamba's ability to comprehend long contexts. ReMamba incorporates selective compression and adaptation techniques within a two-stage re-forward process, incurring minimal additional inference costs overhead. Experimental results on the LongBench and L-Eval benchmarks demonstrate ReMamba's efficacy, improving over the baselines by 3.2 and 1.6 points, respectively, and attaining performance almost on par with same-size transformer models.
\end{abstract}

\section{Introduction}
Transformers~\cite{NIPS2017_3f5ee243}, which form the backbone of most LLMs, encounter substantial challenges when dealing with long texts. The quadratic computational demands and the linear memory costs of the attention mechanism become prohibitive as the text length grows. This complexity poses a significant barrier to effectively modeling long texts, which is crucial for the development of LLMs. To address this, Mamba is proposed as a solution \cite{gu2024mambalineartimesequencemodeling}, which utilizes a recurrent inference mode that ensures linear time complexity and compress information into the fixed state size. This results in constant memory demands during inference. Furthermore, Mamba eliminates the need for positional encoding, theoretically allowing it to handle inputs of any length, while performing competitively against transformers on downstream tasks. Shortly after, Mamba2 was introduced, simplifying the structured $A$ matrix of Mamba to enable faster training and enlarged state size \cite{dao2024transformersssmsgeneralizedmodels}.

\begin{figure}[t]
\centering
\includegraphics[width=0.43 \textwidth]{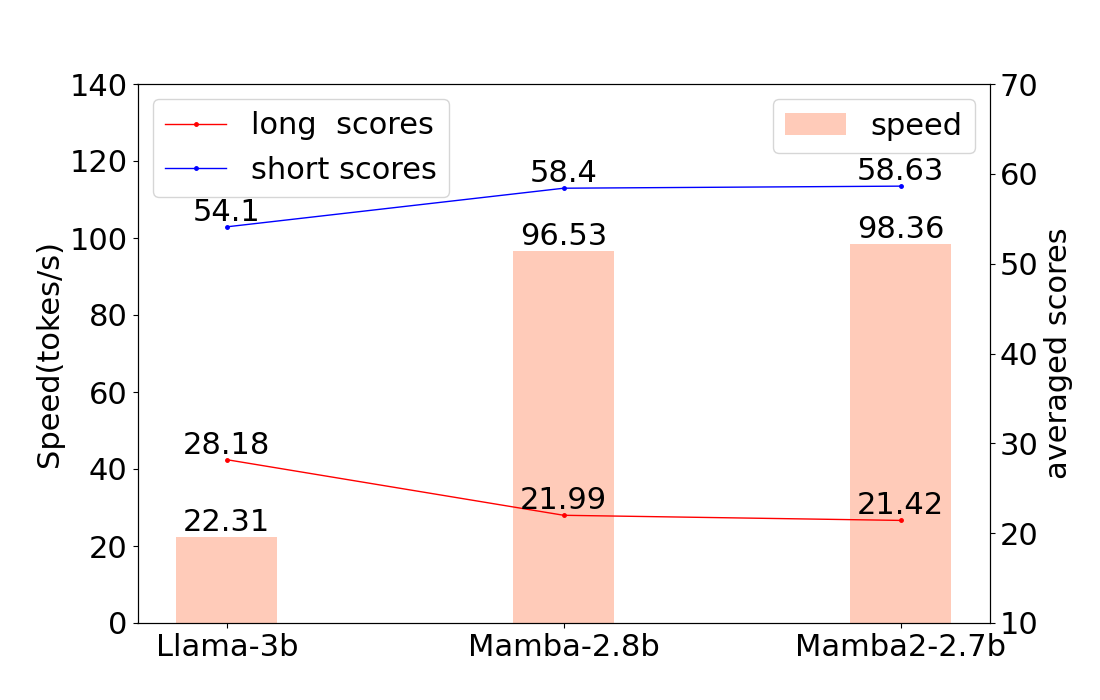} % Reduce the figure size so that it is slightly narrower than the column.
\caption{A comparison of pretrained Mamba models and Transformers of equivalent size across speed, short-context, and long-context performance metrics. Speed is measured under conditions of 6k input tokens and 1k output tokens. ``short scores'' represents the average accuracy across six tasks (HellaSwag, PIQA, Arc-E, Arc-C, WinoGrande, OpenbookQA) evaluated within the LM evaluation harness \cite{eval-harness}. ``long scores'' corresponds to the average scores on the LongBench-E benchmark \cite{bai2024longbenchbilingualmultitaskbenchmark}. Notably, all LongBench evaluations employ a maximum token length of 2k to align with the model's training configuration.}
\label{figreissues}
\end{figure}

Despite these advantages, some studies reveal that Mamba models do not perform as well as expected when dealing with long texts reaching 2k tokens or more~\cite{waleffe2024empiricalstudymambabasedlanguage}. 
As depicted in Figure~\ref{figreissues}, our experimental findings reveal that the pretrained Mamba model surpasses pretrained Transformers of comparable size, such as Llama-3b ~\cite{openlm2023openLlama}, on short-context tasks. Conversely, a substantial performance degradation is observed for Mamba on long-context tasks relative to Transformers.
This performance disparity underscores a significant limitation of Mamba models in practical long-context applications.

This long-context deficency issue of Mamba is usually attributed to its RNN-like nature. This kind of architecture exhibits limitations in preserving crucial information from earlier input sequences as the context length increases due to the fixed-size memory~\cite{wen2024rnns,yang2024efficient}. 
% This forgetting mechanism results in poor performance on long-context tasks, a problem not adequately addressed in existing research on Mamba models. 
Hybrid architectures~\cite{lieber2024jambahybridtransformermambalanguage,ren2024sambasimplehybridstate,park2024mambalearnlearncomparative} have sought to mitigate this issue by integrating attention mechanisms from transformers. However, these approaches often lead to decreased computational efficiency and increased memory consumption. 

The key challenge in Mamba is the excessive degradation of distant information. ReMamba addresses this by employing an effective compression strategy that condenses the information and reduces the context length. Specifically, it selects the top-k hidden states during the first forward pass and integrates them into the state space using Mamba's selective mechanism in the second pass. ReMamba introduces minimal computational overhead (one additional forward pass) and maintains low, constant memory consumption. Experimental results demonstrate that our approach significantly improves Mamba's long-context performance, bringing it close to the performance of transformers. Our ReMamba model achieves a 3.2 improvement over the baseline on LongBench \cite{bai2024longbenchbilingualmultitaskbenchmark} and 1.6 improvement on L-Eval \cite{an2023levalinstitutingstandardizedevaluation}. Furthermore, our methodology exhibits transferability to Mamba2, yielding a 1.6 improvement on LongBench.

\section{Related work}

\subsection{Mamba}
The state space model chooses the time-invariant $\hat{A}$ (state transition matrix) and $\hat{B}$ (input coefficient matrix) thus lacking expressiveness and flexibility. Mamba \cite{gu2024mambalineartimesequencemodeling} proposes to make  $\hat{A}$ and $\hat{B}$ dynamically depend on inputs.

Recall that in one Mamba layer \(l\) , SSM states $S$ are transformed as follows:

\begin{subequations}
\label{e3}
\begin{align}
\label{e3a}
\Delta_{t-1}^{l} &= \mathrm{Softplus}\left(\text{Proj}_{1}(h_{t-1}^{l-1})\right), \\
B_{t-1}^{l} &= \text{Proj}_2\left(h_{t-1}^{l-1}\right), \\
\hat{A}^l, \hat{B}_{t-1}^l &= \text{discretize}\left(A^l, B_{t-1}^l, \Delta_{t-1}^{l}\right), \\
h^{\prime l}_{t-1} &= \text{Proj}_3\left(h_{t-1}^{l-1}\right), \\
S_t^l &= \hat{A}^l \otimes S_{t-1}^l + \hat{B}_{t-1}^l  \left({h^{\prime l}_{t-1}}\right)^T.
\end{align}
\end{subequations}

\noindent Here, $h_{t-1}^{l-1} \in \mathbb{R}^{H}$ represents the output hidden state of Mamba at layer $l-1$ and time step $t-1$. The Softplus function is denoted by $\mathrm{Softplus}$, and $\mathrm{Proj}_1$, $\mathrm{Proj}_2$, and $\mathrm{Proj}_3$ are abbreviations for multiple space projection operations.

Furthermore, $\Delta_{t-1}^{l} \in \mathbb{R}^{H'}$ is the discrete time step corresponding to the selective mechanism in Mamba, where $H'$ is the intermediate hidden size. The continuous and discrete state transformation matrices at layer $l$ are given by $A^l, \hat{A}^l \in \mathbb{R}^{H' \times N}$, respectively. The continuous and discrete input coefficient matrices are denoted by $B_{t-1}^{l}, \hat{B}_{t-1}^l \in \mathbb{R}^{N \times 1}$. The state size is represented by $N$. The discretization method for computing $\hat{A}$ and $\hat{B}$ is indicated by ``discretize''. The vector $h_{t-1}^{'l} \in \mathbb{R}^{H' \times 1}$ and the SSM state is represented by $S_t^l \in \mathbb{R}^{H' \times N}$. The symbol $\otimes$ denotes element-wise multiplication, and $\hat{B}_{t-1}^l \left({h^{\prime l}_{t-1}}\right)^T$ represents matrix multiplication. 

% It is important to note that the definitions of $A^l$ and $\hat{A}^l$ presented here differ from their original definitions due to Mamba's simplification to diagonal matrices.
% The Mamba model is memory-efficient and performs inference in a recurrent mode, enhancing speed. 

% Its recurrent nature also eliminates the need for positional encoding.

\subsection{Mamba2}
\citet{dao2024transformersssmsgeneralizedmodels} theoretically proves the connections between structured state space models and attention mechanisms. They also simplify structured matrix $\hat{A}$ further into scalar-times-identity structure and thus develop a new state space duality (SSD) framework with multi-head patterns similar to transformers.

% \huzhang{You may move the leading sentences and the references of KV cache compression to the place after the introduction of ReMamba. Otherwise readers may not get the point of shared spirits between ReMamba and KV cache compression. }
%
%\subsection{KV cache compression}
%\citet{NEURIPS2023_3d77c6dc} propose a method to compress the past KV cache into gist tokens. \citet{ge2024incontextautoencodercontextcompression} propose to compress the context into memory slots with an additional encoder. \citet{yang2024pyramidinferpyramidkvcache} propose to reduce the KV cache according to the consistency in attention weights.
\subsection{Long Context Mamba and Transformers}
Positional interpolation has been widely used as a technique to extend the context length of transformers \cite{chen2023extendingcontextwindowlarge, yarn,longrope}. But they are specialized for transformers.

Mamba has been found to struggle in maintaining performance beyond its pretraining context length without additional training. LongMamba \cite{longmamba} made the first successful attempt to extend Mamba's context length through a few hours of long-context fine-tuning. DeciMamba \cite{benkish2024decimambaexploringlengthextrapolation} aimed to address the context extension problem of Mamba in a training-free manner, proposing a method to progressively reduce sequence length across layers by empirically removing unimportant tokens.

However, our experiments demonstrate that long-context fine-tuned Mamba still lags behind long-context fine-tuned transformers of the same size, despite using the same data. Moreover, DeciMamba2.8b appears to be insufficiently effective when evaluated on two widely used long-context benchmarks.

\section{Preliminary Study}
KV cache compression is widely used in transformers to reduce memory consumption and improve inference speed. However, prompt compression often results in performance degradation compared to the full context lengths generation in transformers. Unlike transformers, Mamba does not employ a KV cache; instead, it utilizes a fixed-size state space in each layer to preserve context memory. A potential issue with Mamba is its tendency to forget distant information. In our preliminary study, we hypothesize that the state space update in Mamba is insufficient for effectively compressing context information, and that techniques like prompt compression could help relieve this issue.

To explore this, we apply a simple prompt compression method: we replace part of the context tokens with a few randomly selected hidden states from the last layer of Mamba, creating a shorter prompt (referred to as random Mamba). This approach is similar to that of \citet{ge2024incontextautoencodercontextcompression}, which used soft prompts for information compression. Intuitively, this random compression method may lead to significant information loss and degraded performance. However, our results show that the average scores for different context lengths on LongBench between normal Mamba and random Mamba are similar when both are trained on the same long-context dataset. Furthermore, random Mamba outperforms normal Mamba at certain context lengths, as shown in Figure \ref{figalbation} of \ref{ablation}. The random\_select (SFT) represents the fine-tuned random Mamba, while Mamba (SFT) represents the fine-tuned vanilla Mamba.

This observation suggests that information loss in Mamba when handling long contexts is substantial. To relieve this, we propose selective compression and selective adaptation through leveraging Mamba’s state space update mechanism.

\begin{figure*}[t]
\centering
\includegraphics[width=0.8 \textwidth]{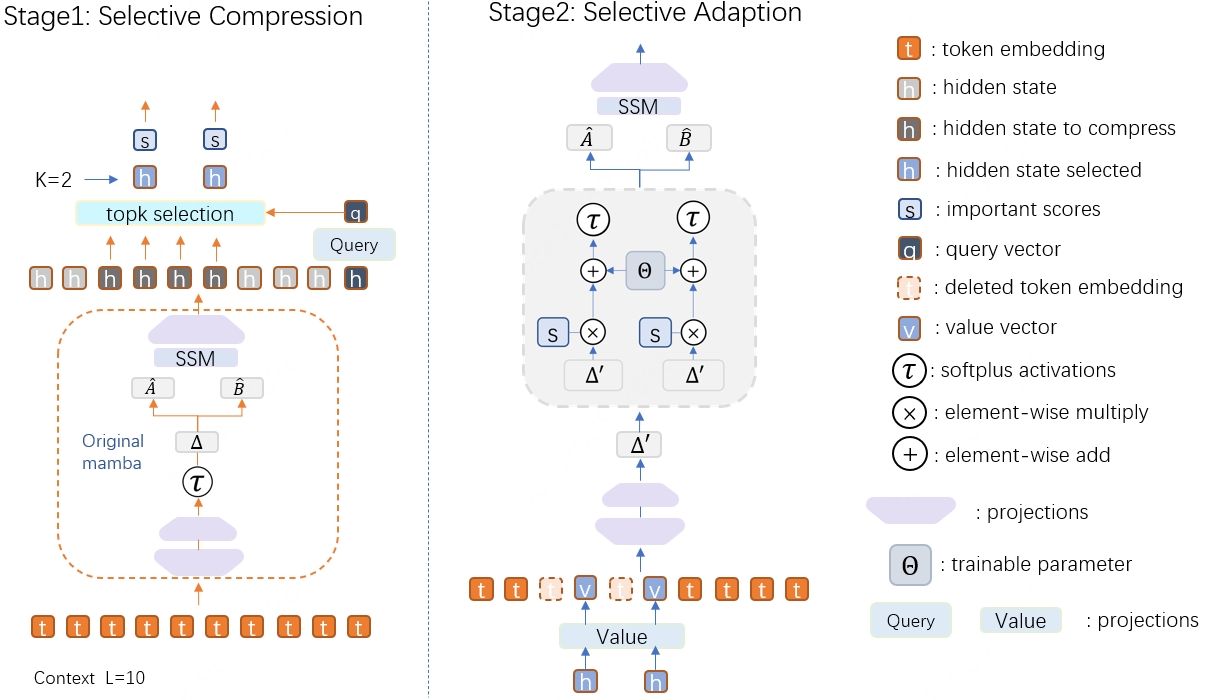} % Reduce the figure size so that it is slightly narrower than the column.
\caption{ReMamba architecture. We just show one layer and leave out the $A$, $B$ and discrete method here. For Stage 2, only those value vectors selected need to go through selective adaption. Normal token embeddings just flow as usual. We select top-$K$ (here is top-2) hidden states in the last layer according to their importance scores calculated with the last hidden state $h_L$. And we incorporate the scores into the gradient utilizing the selective mechanism in Mamba.}
\label{figremamba}
\end{figure*}

\section{Methodology}

% \bei{In this work, we propose a Remamba model that could alleviate the performance degradation for Mamba when facing a long input sequence, e.g., extending 2K input tokens. Our method comprises twice forward evaluations, and we will elaborate on the details as follows.}

% In this work, we propose a ReMamba model that can alleviate the performance degradation for Mamba when facing a long input sequence. We explicitly select and compress the key information in the distant and reduce state space update to relieve the sever information loss. Our method comprises two forward stages.\
ReMamba consists of two forward stages. In the first stage, three feed-forward networks are employed to help determine the significance of hidden states from Mamba's final layer. These hidden states are selected based on their importance scores.
The second stage integrates these compression hidden states with the input context, adapting Mamba's selective mechanism to incorporate them into the state space. 

Our proposed method draws some spirits from techniques employed in KV cache compression \cite{NEURIPS2023_3d77c6dc,ge2024incontextautoencodercontextcompression,yang2024pyramidinferpyramidkvcache,DBLP:conf/emnlp/ChevalierWAC23,hwang2024transformerfamfeedbackattentionworking,gao2024selfcpcompressingoverlimitprompt} by leveraging the language model itself to aggregate information via hidden states and employing a scoring mechanism to select the most salient representations. Nevertheless, different from transformers, ReMamba's compression strategy focuses on two key objectives: 1) compressing and selectively retaining crucial information to minimize information degradation, and 2) reducing the frequency of state space updates to further alleviate the information loss. 

The selection method employed in ReMamba is a simplified key-query-value-based approach, commonly used in Retrieval-Augmented Generation and summarization tasks \cite{DBLP:conf/nips/LewisPPPKGKLYR020, mao-etal-2022-dyle}. In this context, we select the most important hidden states from the last layer of Mamba to mitigate information loss during the compression process.

\subsection{Stage1 : Selective Compression}
Selective compression involves selectively compressing the input prompt by leveraging the final layer hidden states of the Mamba model to decrease state updates and consolidate information.

% Suppose that the sequence length is $L$ and the context token embeddings are $\{t_i\}_{i=1}^{L}$, we first define the relative range to compress as $range:=(s, e), e=s+p$ where  $s$ and $e$ are the relative start and end positions,  and $p$ is the relative length to compress, satisfying $0\le s, p, e\le 1$. The index set of the context to compress is $\mathcal{R} :=[S, E]$ where $S=L*s+1, E=L*(s+p)$. Therefore, the length of the prompt to compress is $L^{'}=E-S+1$. We abuse $\mathcal{R}$ to represent the set of indices and the set of actual tokens of the context to compress. We further define the compression ratio $\rho$, and we  compress the selected context $\mathcal{R}$ into $K := |\mathcal{R}|*\rho$ hiddens.
Suppose the sequence length is $L$ and the context token embeddings are $\{t_i\}_{i=1}^L$. We define the relative range to be compressed as $range := (s, e)$, where $e = s + p$, with $s$ and $e$ denoting the relative start and end positions, respectively, and $p$ representing the relative length to compress. These values satisfy $0 \leq s, p, e \leq 1$. The index set of the context to compress is $\mathcal{R} := [S, E]$, where $S = L \cdot s + 1$ and $E = L \cdot (s + p)$. Consequently, the length of the prompt to compress is $L' = E - S + 1$. For convenience, we use $\mathcal{R}$ to represent both the set of indices and the set of actual tokens within the context to be compressed. Furthermore, we define the compression ratio $\rho$ and compress the selected context $\mathcal{R}$ into $K := |\mathcal{R}| \cdot \rho$ hidden representations.

% In Figure \ref{figremamba}, the compression hyperparameter setting is:
% $s=0.2, p=0.4, range:=(0.2, 0.6) , \mathcal{R} :=[3, 6] , \rho=0.5, K=2$ .
% In our experiment, we find that $s=0$ is the best, which is due to the casual language modeling nature of Mamba (we will discuss this later).
In Figure~\ref{figremamba}, the compression hyperparameter settings are:
$s = 0.2$, $p = 0.4$, $range = (0.2, 0.6)$, $\mathcal{R} = [3, 6]$, $\rho = 0.5$, $K = 2$.
In our experiments, we find that $s = 0$ yields the best results, which can be attributed to the casual language modeling nature of Mamba (this will be discussed in more details in Appendix \ref{sec:appendixs}).

As shown in the Stage 1 of Figure \ref{figremamba}, we denote the last layer's output hidden states as $\{h_i\}_{i=1}^L$, where each $h_i \in \mathbb{R}^H$  with $H$ representing the hidden size.
We then transform the last hidden state $h_L$ into a query hidden state, namely $q$, through a feed-forward layer named $Query$. Additionally, the hidden states to be compressed, denoted as $\{h_i\}_{i=S}^{E}$, are transformed into $\{k_i\}_{i=S}^{E}$ via a $Key$ layer (this transformation is not shown in Figure \ref{figremamba}). Finally, the cosine similarity scores, \(Cos=\{cos_i\}_{i=S}^{E}\), are computed to serve as importance scores for the hidden states $\{h_i\}_{i=S}^{E}$. 
The calculation of $q$, $k_i$, and $cos_i$ is formulated as follows:
\begin{equation}
\begin{aligned}
q &= \mathrm{Query}(h_{L}) \\
\{k_i\}_{i=S}^{E} &= \mathrm{Key}(\{h_i\}_{i=S}^{E}) \\
cos_i &= \frac{k_i \cdot q}{\max(\Vert k_i \Vert_2 \cdot \Vert q \Vert_2, \epsilon)}
\end{aligned}
\end{equation}
\noindent where $k_i$ represents the transformed hidden state at position $i$, and $cos_i$ computes the cosine similarity between $q$ and $k_i$. The constant $\epsilon$ prevents division by zero.

We select the top-$K$ hidden states ${h_j}$, where $j \in G$, from the hidden states $\{h_i\}_{i=S}^{E}$ based on their importance scores, denoted by $Cos$. The index set $G$ is defined as:

\begin{equation}
G = \mathop{\arg\max}_{A \subset \{S, S+1, ..., E\}, |A| = K} \sum_{i \in A} cos_i
\end{equation}

Note that the original order of these indices is preserved.

In our model, after selecting the top-$K$ hidden states $h_j$, we apply a feed-forward layer, $Value$, to project them into the token embedding hidden space:

\begin{equation}
\{v_i\}_{i=1}^{K} = V(\{h_j\}, j \in G)
\end{equation}

Their corresponding cosine similarity scores are $\{cos_i'\}_{i=1}^{K}$. We then replace the token embeddings $\{t_i\}_{i=S}^{E}$ ($\mathcal{R}$) with $\{v_i\}_{i=1}^{K}$. Consequently, the new input embeddings for Mamba are replaced by:

\begin{align}
T_{\text{new}} &= \text{Cat}(\{t_i\}_{i=1}^{S-1}, \{v_i\}_{i=1}^{K}, \{t_i\}_{i=E+1}^{L}) \\
&= \{t_i'\}_{i=1}^{L-L'+K}
\end{align}

\noindent where $\text{Cat}$ denotes the concatenation operation. The length of $T_{\text{new}}$ is $L-L'+K$, resulting in a significantly shorter input sequence for the second forward pass compared to the first.

\subsection{Stage 2: Selective Adaption}

One significant challenge in using top-$K$ selection based on importance scores is its non-differentiability, which impedes the ability to train such models effectively.
Here we propose a framework that integrates importance scores into the selective mechanisms of the Mamba model.

% Common approaches mitigate this by directly multiplying the hidden states with their corresponding scores \cite{raposo2024mixtureofdepth}. While our preliminary experiments confirm the effectiveness of this method, we propose an enhanced framework that integrates importance scores into the selection mechanisms of the Mamba model more seamlessly. \huzhang{this sentence does not convey much information. like why it is "more seamlessly"?  I suggest that we do not mention any previous approach because no one has proposed ReMamba structure before. Directly state how we do. } This new approach facilitates a natural gradient flow. 
% \huzhang{"natural gradient flow" has its own meaning. do not use this expression here.}
% \paragraph{Mamba}

For hidden states (embeddings) that do not require compression in stage 1, namely $\{t_i\}_{i=1}^{S-1}$ and $\{t_i\}_{i=E+1}^{L}$, the standard Mamba algorithm is applied during the second forward pass. For embeddings at selected positions, specifically $\{t_{i}^{'}\}_{i=S}^{S+K-1}$ or equivalently $\{v_i\}_{i=1}^{K}$, Equation \ref{e3a} is reformulated as follows:

\begin{equation}
\begin{aligned}
\alpha &= \mathrm{ReLU}(cos_{t-1}^{'}) \\
{\Delta_{t-1}^{l}}^{'} &= \text{Proj}_{1}(h_{t-1}^{l-1}) \\
\delta &= {\Delta_{t-1}^{l}}^{'} \cdot \alpha + \Theta^{l} \\
\Delta_{t-1}^{l} &= \mathrm{Softplus}(\delta)
\end{aligned}
\end{equation}

\noindent where $\Theta^{l} \in \mathbb{R}^{H'}$ is a layer-wise trainable offset parameter controlling scale intensity. $\mathrm{ReLU}$ is the activation function. Intuitively, hidden states with low importance scores should minimally impact model computations. Therefore, we approximate this behavior by setting their corresponding $\Delta$ values close to zero. Ideally, directly multiplying $\Delta$ by $\alpha$ would be more precise, but this necessitates modifications to the selective scan algorithm, leading us to adopt the simpler approach.

% For those hidden states (embeddings) that do not need to be compressed in stage 1 ($\{t_i\}_{i=1}^{S-1}$ and $\{t_i\}_{i=E+1}^{L}$), they just go through normal Mamba algorithms in the second foward pass.
% For those embeddings in selected position ($\{t_{i}^{'}\}_{i=S}^{S+K-1}$ that is , $\{v_i\}_{i=1}^{K}$) , 
% equation \ref{e3a} is reformulated as follows:
% %  
% \begin{equation}
% \begin{aligned}
% \alpha &= \mathrm{ReLU}(cos_{t-1}^{'}) \\
% \Delta_{t-1}^{l}^{'} &= \text{Proj}_{1}(h_{t-1}^{l-1}) \\
% \delta &= \Delta_{t-1}^{l}^{'}\cdot \alpha + \Theta^{l} \\
% \Delta_{t-1}^{l} &= \mathrm{Softplus}(\delta) 
% \\
% \end{aligned}
% \end{equation}
% %   
% \noindent where $\Theta^{l} \in \mathbb{R}^{H^'}$ is  a trainable offset parameter layer-wise to control scale intensity. $\mathrm{ReLU}$ is relu activation function. The intuition is straightforward: hiddens with low importance scores should have small impact on model's computations. So we approximate it by setting their corresponding $\Delta$ values close to zero. Ideally, a more precise approach is to directly multiplying $\Delta$ with $\alpha$. However, implementing this requires modifications to the selective scan algorithm, prompting us to adopt the simpler alternative. 

\subsection{Training}
Following the forward encoding processes, standard causal language generation is applied using the Mamba architecture. During training, newly introduced parameters within the selective compression mechanism are optimized. These parameters, except for $\Theta$ which is initialized to all zeros, are initialized with a subset of the weights from the first layer's $\textrm{in\_proj}$ matrix. Additionally, for parameters in Mamba, the $\text{dt\_proj}$ matrix is fully trained, while $\text{in\_proj}$, $\text{out\_proj}$, $\text{embeddings}$, and $\text{lm\_head}$ are updated using Low-Rank Adaptation (LoRA) \cite{DBLP:conf/iclr/HuSWALWWC22}. In our best implementation, to emphasize the significance of specific information, gradients flowing into the importance scores are scaled proportionally to these scores. This approach intuitively prioritizes the training of more critical representations.

% After this two forward pass encoding process, causal language generation is applied as usual Mamba.
% During training, we train new parameters introduced by our selective compression mechanism. We initialize them with part of the weights of the $\textrm{in\_proj}$ in the first layer. We also train full $\text{dt\_proj}$ and train $\text{in\_proj}$, $\text{out\_proj}$, $\text{embeddings}$, $\text{lm\_head}$ with LoRA. In our best implementation, the gradients flow into the important scores are rescaled by the important scores itself. An intuition here is that for those with higher important scores, their information counts more. So it is better to train more.
% \huzhang{These sentences are toooo casual. Be formal and accurate.}
% \subsection{Time complexity analysis}
% Here we analyse the time complexity of ReMamba.
% Supposing sequence length is $N$, hidden dimension is $H$. Th

%----------------------------------------------

\newcommand{\M}{50}
\begin{table*}[t]
    
  \setlength{\tabcolsep}{0.8pt}
  \small
  \centering
  \begin{adjustbox}{width=2.05\columnwidth,center}
   \begin{tabular}{lcccccccccccccc}

  \toprule
  
  \bf Model & \bf \rotatebox{\M}{2WikiMQA}  & \bf \rotatebox{\M}{GovReport} & \bf \rotatebox{\M}{HotpotQA}  & \bf \rotatebox{\M}{LCC}  
  & \bf \rotatebox{\M}{MultiNews} & \bf \rotatebox{\M}{MultiQA} & \bf \rotatebox{\M}{PassCount}  & \bf \rotatebox{\M}{PassRetrie.}  & \bf \rotatebox{\M}{Qasper} & \bf \rotatebox{\M}{RepoBench} & \bf \rotatebox{\M}{SAMSum} & \bf \rotatebox{\M}{TREC} & \bf \rotatebox{\M}{TriviaQA} & \bf \rotatebox{\M}{Average}  \\
  \midrule
  Llama-3b~(Pre)  & 4.07& 10.33& 3.93& 43.17& 6.96& 7.74& 1.78& 7.13& 3.49& 32.14& 9.98& 15.0& 30.65& 13.57       \\

  Llama-3b~(SFT)    & 15.69& 24.55& 19.69& 56.17& 19.37& 29.73& 0.33& 6.67& 19.73& 44.47& 33.37& 48.67& 58.41& 28.99 \\    

\midrule
  DeciMamba~(Pre)& 3.89& 9.36& 3.85& 25.69& 11.62& 4.91& 0.83& 1.19& 3.21& 13.88& 6.79& 6.67& 17.72& 8.43 \\
  DeciMamba~(SFT)& 19.6& 13.74& 15.32& 23.65& 10.91& 17.88& 1.37& 4.96& 5.72& 12.54& 10.59& 30.33& 46.06& 16.36 \\

  Mamba~(Pre)& 3.73& 8.72& 4.03& 24.03& 11.31& 4.95& 0.80& 1.75& 3.67& 12.83& 6.86& 9.00& 17.40& 8.39 \\
   
  Mamba~(SFT)  & 22.10& 19.08& 15.90& 40.20& 19.36& 30.28& 0.00& 4.67& 19.04& 36.02& 28.30& 39.33& 45.97& 24.63                   \\ 
  ReMamba~(SFT)    & 21.18& 19.67& 20.56& 48.21& 18.86& 26.39& 3.21& 6.83& 16.76& 40.40& 33.65& 48.67& 57.73& 27.86            \\
  \bottomrule
  \end{tabular}
  \end{adjustbox}

\caption{Performance on LongBench-E (English branch). ``MultiQA'' denotes MultiFieldQA , ``PassCount'' denotes PassageCount, ``PassRetrie.'' denotes PassageRetrieval.
 Models are evaluated using a maximum length of 6K tokens, matching their finetuning configurations. 
% This 2K setting yields optimal performance in our experiments. 
Here ``(Pre)'' means pretrained model. ``(SFT)'' means finetuned model. 
% All finetuned models, except Llama-2-3B, are trained on the same dataset. Llama-2-3B's finetuning process also uses the same dataset but truncates input sequences to 2K tokens during training, which proves to be the best configuration.
}
\label{tab:longbench}
\end{table*}

%----------------------------------------------

\begin{table*}[t]
    
  \setlength{\tabcolsep}{5.5pt}
  \small
  \centering

  \begin{tabular}{lcccccccccccccc}

  \toprule
  
  \bf Model & \bf Finetuned & \bf Tokens & \bf CodeU  & \bf Coursera & \bf GSM  & \bf QuALITY 
  & \bf SFictio  & \bf TOEFL  & \bf Average  \\
  \midrule
  Llama-3b~(Pre) & \XSolidBrush  & 6k & 0.0& 24.71& 3.0& 27.23& 32.81& 17.47& 17.54     \\
  Llama-3b~(SFT)  & \Checkmark   & 6k & 1.11& 19.62& 7.0& 24.26& 57.03& 27.14& 22.69    \\
   
  \midrule
   DeciMamba~(Pre) & \XSolidBrush & 6k  & 2.22& 24.71& 0.0& 25.25& 24.6& 16.73& 15.59 \\
    DeciMamba~(SFT) & \Checkmark & 6k & 0.0& 23.69& 0.0& 26.86& 53.17& 22.68& 21.07   \\ 
  
  Mamba~(Pre) & \XSolidBrush & 6k  & 2.22& 23.26& 0.00& 25.74& 23.44 & 17.10& 15.29      \\   
  
  Mamba~(SFT) & \Checkmark & 6k  & 4.44& 26.16& 1.00& 27.72& 50.78& 23.05& 22.19   \\ 
  ReMamba~(SFT)  & \Checkmark & 6k & 2.22& 22.97& 3.00& 25.74& 58.59 & 30.48& 23.83  \\

  \bottomrule
  \end{tabular}
  
  \caption{Model performance on closed-ended tasks of L-Eval. ``Tokens'' denotes the max length. ``SFT'' denotes finetuned models. ``Pre`` denotes pretrained models.
  % Llama-2-3B's finetuning process uses the same dataset as Mamba and ReMamba but truncates input sequences to 2K tokens during training, which proves to be the best configuration.
  }
  \label{tab:lEvalVaringL}
\end{table*}

\section{Experiments}
\subsection{Experimental Setups}
Our model is designed for long-context question-answering tasks, necessitating a substantial corpus of long-context instruction tuning data. To this end, we leverage the OpenOrca dataset \cite{mukherjee2023orcaprogressivelearningcomplex} and LongAlpaca-12k \cite{DBLP:conf/iclr/ChenQTLL0J24}. The former comprises a rich collection of ChatGPT-augmented FLAN data alignments, while the latter is a long-context alignment dataset. We initially filter long instruction tuning instances from OpenOrca and concatenate them with LongAlpaca. To accommodate device memory constraints, prompts are truncated to a maximum length of 6,000 tokens. This process yields approximately 200,000 long-context training examples. To augment training data diversity, the initial 300,000 standard instances from OpenOrca are incorporated. This dataset is referred to as the LongOrca dataset.
We finetune the baseline Mamaba 2.8b model and our ReMamba model on the same dataset. We also finetune a DeciMamba2.8b~\cite{benkish2024decimambaexploringlengthextrapolation} and a Llama-3b~\cite{openlm2023openLlama} for reference. DeciMamba aims to address Mamba's context-extension issue without requiring additional training. Although our approach differs slightly in terms of settings and objectives, we still fine-tune 
DeciMamba2.8b using the same data. Given the 2k maximum positional encoding limit of Llama-3b, we conduct fine-tuning experiments using the simple linear positional interpolation technique \cite{chen2023extendingcontextwindowlarge} to extend its context length. The data construction process for Llama-3b is identical to that of Mamba. Details can be found in \ref{sec:appendixdetail}.
Notice that Mamba2.8b is the largest model we can obtain.

\subsection{Evaluations}
We conduct comparative analyses of our model against baseline Mamba2.8b and DeciMamba2.8b (both of finetuned and pretrained) on the widely adopted LongBench benchmark~\cite{bai2024longbenchbilingualmultitaskbenchmark} and LEval benchmark~\cite{an2023levalinstitutingstandardizedevaluation}, which encompass a diverse set of challenging real-world long-context tasks. For consistency, the same prompt templates and greedy decoding configurations are employed across all models.

To provide a reference point, the performance of a similarly sized transformer architecture (Llama-3b) is also included. Both the pretrained and fine-tuned evaluations of Llama-3b utilize the linear positional interpolation technique.
% We also test the inference expense of our model comparing to the baseline Mamba and other models.

\subsection{Results}

\paragraph{Results on LongBench}
% Here we compare the model performance on the LongBench english branch for our training set only contains english.
% As shown in fig \ref{figrelongbenchVaringL}, We vary the maximum prompt length ranging from 2k to 9k and record the average scores on remamba, finetuned mamba and pretrained mamba. The curve shows that our method not only achieve higher scores on various length but also extend the efficient length of Mamba model. Scores show no much degradation to a relative longer context length.

We choose the English branch of LongBench because our training set only contains English. Higher values across all indicators are indicative of better performance.
We compare the performance of the models in detailed tasks in Table \ref{tab:longbench} under the max length 6k corresponding to the training setting. Here the hyperparameters for ReMamba are: $s=0$, $p=0.18$ and $\rho=0.009$. We will also show later that our model's robustness to various of hyperparameter combinations.
Table \ref{tab:longbench} shows that our ReMamba model improves the average scores on LongBench 3.23 compared to the SFT Mamba baseline. Our model approaches the pretrained and finetuned transformer baseline. The results for DeciMamba indicate that it may not be sufficiently effective for tasks in LongBench or it may be sensitive to the choosing of hyperparameters.

\paragraph{Results on LEval}
We compare the performance on the closed-ended tasks of L-Eval. The higher all indicators are, the better.
A snap of detailed task scores for the maximum length of 6k is presented in Table \ref{tab:lEvalVaringL}. We can witness a 1.64 improvement on average scores compared to the SFT Mamba baseline. Here the hyperparameter setting for ReMamba is: $s=0$,
$p=0.20$ and $\rho=0.05$. The results for DeciMamba2.8b also show no significant improvements.

\subsection{Analyses and Discussions}

\subsubsection{Ablation Study}
\label{ablation}
To verify the effectiveness of the modules we introduced, we conduct an ablation study by comparing ReMamba against three alternative methods: 
% To verify the effectiveness of all the modules we introduce, we demonstrate three reduced methods including 
1. Random Selection: which randomly select hidden states as the compressed information according to $\rho$. 2. Fix Selection: given the $\rho$ we select enough hidden states every $k$ positions. 
% Here $k$ is an interval that can be calculated. 
The interval 
$k$ is calculated based on the compression ratio. 
3. Multiplicative Selection: This variant just modifies the selective adaptation process by directly multiplying importance scores with the selected hidden states, aligning with the approach proposed by \citet{raposo2024mixtureofdepth}. 
All of those models are trained on the same data as ReMamba.

\begin{figure}[t]
\centering
\includegraphics[width=0.5\textwidth]{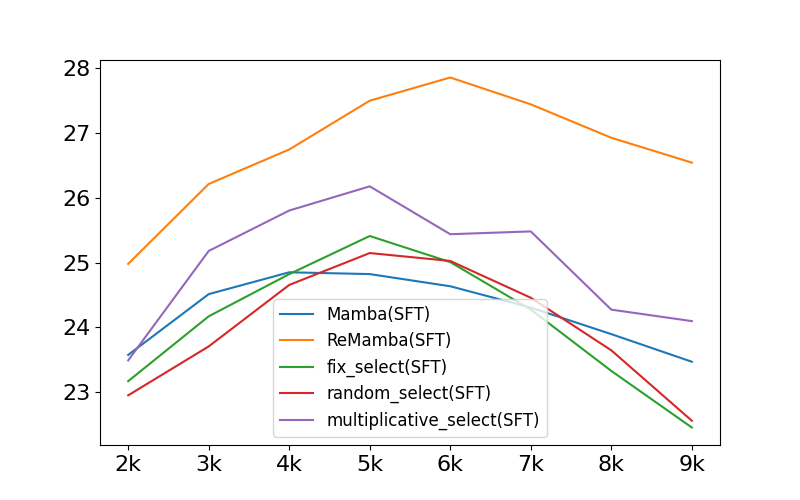} % Reduce the figure size so that it is slightly narrower than the column.
\caption{Ablation study about average scores on LongBench varying max length from 2k to 9k. ``Mamba(SFT)'' is the finetuned Mamba. ``fix\_select'' is the Fix Selection. ``random\_select'' is the Random Selection. ``multiplicative\_select'' is the Multiplicative Selection.}
\label{figalbation}
\end{figure}

We report the averaged scores on LongBench across various maximum input lengths. As illustrated in Figure \ref{figalbation}, both the fixed and random selection methods achieve performance comparable to the finetuned Mamba baseline. Interestingly, these methods even outperform Mamba at lengths of 5k and 6k. This observation confirms our hypothesis that Mamba models suffer from severe forgetting issues. Even simple methods like dropping some information appear beneficial. The performance of the multiplicative selection method shows some improvements across varying input lengths. However, the substantial performance gap observed with our selective adaptation module demonstrates its critical role in the ReMamba model. The selective adaptation module not only mitigates the forgetting problem, but also significantly enhances the model's ability to handle longer input sequences effectively.

\begin{figure}[t]
\centering
\includegraphics[width=0.5 \textwidth]{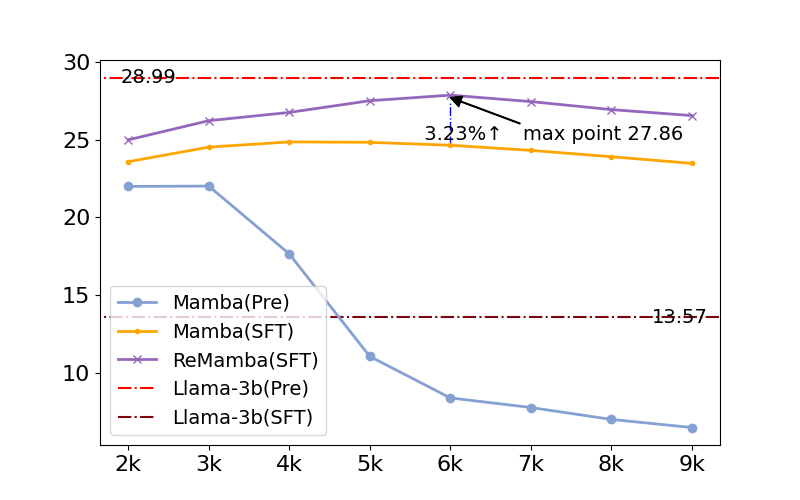} 
\caption{Average scores on LongBench varying max length from 2k to 9k. 
The ``Pre'' means pretrained model while ``SFT'' means finetuned model.
The performance of Llama-3b~(SFT) and Llama-3b~(Pre) is for reference, using the max length of 6k.
% Mamba~(pretrain) means Mamba pretrained model, Mamba~(SFT) means Mamba finetuned model, ReMamba~(SFT) means ReMamba model. 
}
\label{figrelongbenchVaringL}
\end{figure}

\begin{figure}[t]
\centering
\includegraphics[width=0.5\textwidth]{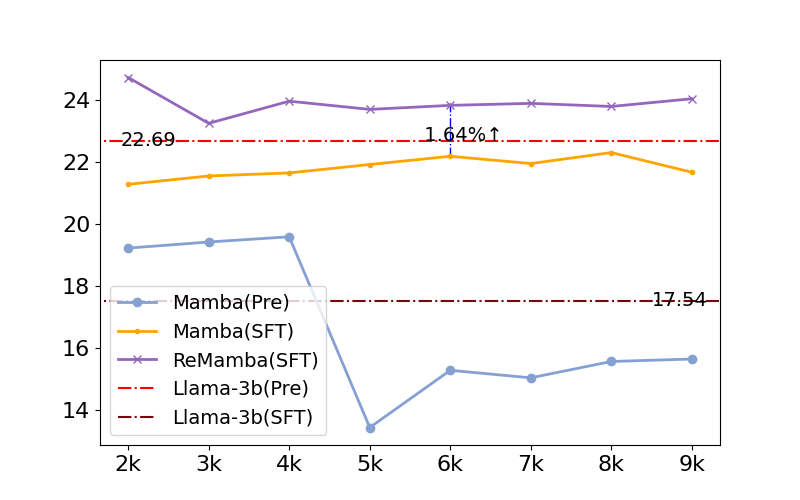} % Reduce the figure size so that it is slightly narrower than the column.
\caption{Average scores on L-Eval varying max length from 2k to 9k. The performance of Llama-3b (SFT) and Llama-3b (Pre) is for reference, using the max length of 6k.
}
\label{figrelEvalVaringL}
\end{figure}

\begin{figure}[t]
\centering
\includegraphics[width=0.40\textwidth]{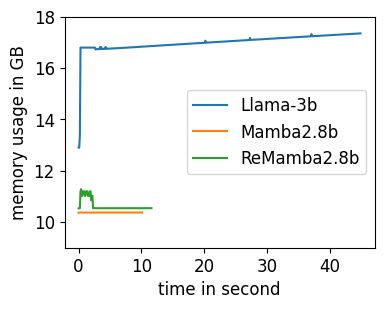} % Reduce the figure size so that it is slightly narrower than the column.
\caption{Memory consumption comparisons during inference are conducted with a 6144 input sequence and a 1024 output sequence, using a batch size of 1. The experiments are done on an A100 80GB GPU.}
\label{figrememory}
\end{figure}

\subsubsection{Varying Length}
To complement our main results, which employ a maximum sequence length of 6k tokens to align with training settings, we further evaluate the model performance at varying input lengths ranging from 2k to 9k tokens. This evaluation is conducted using the LongBench and L-Eval benchmarks. As depicted in Figure \ref{figrelongbenchVaringL}, our ReMamba consistently outperforms the baseline Mamba model across all tested context lengths on LongBench. Notably, the performance gap between our model and the baseline widens as the context length increases. Furthermore, our model extends the efficient context length (the length at which greatest performance is observed) to 6k tokens, compared to 4k tokens for the finetuned Mamba baseline.
In Figure \ref{figrelEvalVaringL}, we observe performance improvements across all context lengths for our model on L-Eval.

% In figure \ref{figrelEvalVaringL}, we can also witness performance improvement across all lengths. The averaged scores on L-Eval show a degradation for all models here, this may due to the difficulty of tasks. 
% Note that we can also see our model approaches Llama-3b.

\begin{table*}[t]
    
  \setlength{\tabcolsep}{1.0pt}
  \small
  \centering
  \begin{adjustbox}{width=2.05\columnwidth,center}
  \begin{tabular}{lcccccccccccccc}

  \toprule

  \bf Model & \bf \rotatebox{\M}{2WikiMQA}  & \bf \rotatebox{\M}{GovReport} & \bf \rotatebox{\M}{HotpotQA}  & \bf \rotatebox{\M}{LCC}  
  & \bf \rotatebox{\M}{MultiNews} & \bf \rotatebox{\M}{MultiQA} & \bf \rotatebox{\M}{PassCount}  & \bf \rotatebox{\M}{PassRetrie.}  & \bf \rotatebox{\M}{Qasper} & \bf \rotatebox{\M}{RepoBench} & \bf \rotatebox{\M}{SAMSum} & \bf \rotatebox{\M}{TREC} & \bf \rotatebox{\M}{TriviaQA} & \bf \rotatebox{\M}{Average}  \\

\midrule
  Mamba2~(Pre) & 2.18& 4.76& 1.54& 23.46& 7.71& 2.88& 0.60& 1.47& 1.17& 14.97& 2.07& 8.33& 10.60& 6.29 \\
    
  Mamba2~(SFT)  & 13.73& 19.97& 15.05& 41.78& 19.52& 25.20& 0.67& 5.67& 13.44& 38.79& 33.95& 49.67& 43.54& 24.69                      \\ 
  ReMamba2~(SFT)  & 18.90& 19.03& 18.09& 51.15& 17.68& 25.82& 3.33& 5.00& 14.84& 43.99& 23.55& 44.00& 56.90& 26.33 \\

  \bottomrule
  \end{tabular}
  \end{adjustbox}

\caption{The performance comparisons of LongBench-E (English Branch) on Mamba2. Mamba2~(Pre) means pretrained Mamba2. Mamba2~(SFT) means finetuned Mamba2. ReMamba2~(SFT) means our model. All use the setting of 6k max length.}
\label{tab:longbenchmamba2}
\end{table*}

\subsubsection{Speed Performance and Memory Expense}

Our model introduces a single additional forward pass during inference, resulting in small constant memory consumption. We visualize the memory consumption during inference with a 6k input sequence and a 1k output sequence, using a batch size of 1 in Figure \ref{figrememory}. The device we use is an A100 80GB GPU. We observe that the encoding process of Llama consumes a substantial amount of memory and the KV cache increases gradually during the decoding process, whereas ReMamba incurs only a small additional memory cost. After the encoding process, ReMamba's memory consumption stabilizes at a constant level, which is moderately higher than Mamba's, corresponding to the additional parameters introduced to support the selection mechanism.

\begin{figure}[t]
\centering
\includegraphics[width=0.45 \textwidth]{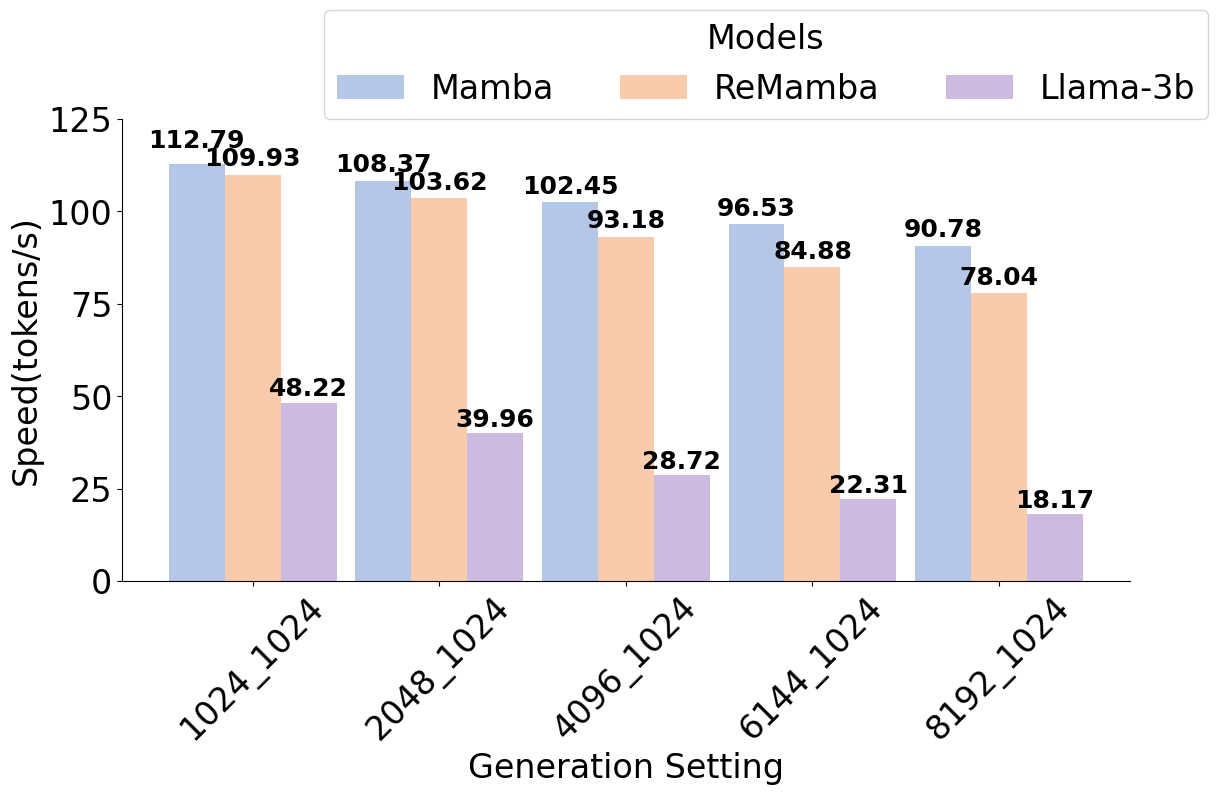} % Reduce the figure size so that it is slightly narrower than the column.
\caption{Speed (tokens/second) performance comparisons. Here 1024\_1024 means input 1024 tokens and output 1024 tokens.}
\label{figrespeed}
\end{figure}

To evaluate the speed performance, we varys the input sequence length from 1k to 8k tokens while fixing the output length at 1k tokens. For all the experiments, we use a batch size of 1 and measure the speed on an NVIDIA A100 80GB GPU. We compare the performance of ReMamba, Mamba, and the vanilla transformer model (Llama-3b), as illustrated in Figure \ref{figrespeed}. The speed metric is given in tokens per second. Our experiments indicate that ReMamba operates at speeds comparable to the original baseline, maintaining a significant speed advantage over traditional transformers.

% Note that for Mamba2, there is no $\textrm{dt\_proj}$ so we train $\textrm{in\_proj}$ and $\textrm{out\_proj}$ with LoRA.

\subsubsection{Robustness varying choices of hyperparamters}
\begin{figure}[t]
\centering
\includegraphics[width=0.38 \textwidth]{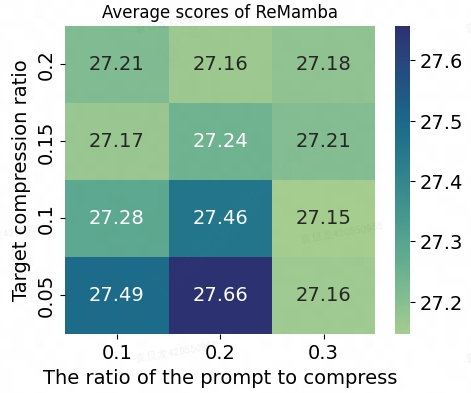} % Reduce the figure size so that it is slightly narrower than the column.
\caption{ Robustness of the ReMamba model with varying hyperparameters. The row label denotes the relative ratio of the prompt to be compressed, corresponding to parameter $p$. The column label indicates the compression ratio, corresponding to parameter $\rho$. }
\label{figueRobust}
\end{figure}

The aforementioned results were obtained using the hyperparameter settings $s = 0$, $p = 0.18$, and $\rho = 0.009$, which demonstrates relatively superior performance. In Figure \ref{figueRobust}, we also show the stability of our model by varying the hyperparameters $p$ and $\rho$. For these experiments, the hyperparameter $s$ is fixed at 0.

\subsubsection{Generalizing to Mamba2}
We also evaluated the applicability of our approach to Mamba2. As shown in Table~\ref{tab:longbenchmamba2}, ReMamba achieves an average improvement of 1.6 points on LongBench. Further details are available in Appendix~\ref{sec:appendixmamba2detail}.

\section{Conclusions}
This study investigates the long-context efficiency challenges posed by Mamba models, hypothesizing that distant information within these models is subject to substantial degradation. In response, we introduce ReMamba, a novel approach that compresses and selectively preserves critical information during an initial forward pass. This compressed information is subsequently integrated into the state space during a second forward pass, capitalizing on Mamba's inherent selective mechanism. Notably, ReMamba incurs minimal computational overhead while substantially enhancing Mamba's long-context performance, thereby offering a promising avenue for advancing the Mamba model family.

\clearpage

\section*{Limitations}

Although ReMamba improves the long-context performance of Mamba, it is unlikely that Mamba can outperform transformers as the context length increases, due to its fixed-size state space. Additionally, ReMamba primarily relieve the loss of long-context information in Mamba through selective compression. A more promising approach would be to directly modify the state space update mechanism.

% \section*{Acknowledgments}

% This document has been adapted
% by Steven Bethard, Ryan Cotterell and Rui Yan
% from the instructions for earlier ACL and NAACL proceedings, including those for
% ACL 2019 by Douwe Kiela and Ivan Vuli\'{c},
% NAACL 2019 by Stephanie Lukin and Alla Roskovskaya,
% ACL 2018 by Shay Cohen, Kevin Gimpel, and Wei Lu,
% NAACL 2018 by Margaret Mitchell and Stephanie Lukin,
% Bib\TeX{} suggestions for (NA)ACL 2017/2018 from Jason Eisner,
% ACL 2017 by Dan Gildea and Min-Yen Kan,
% NAACL 2017 by Margaret Mitchell,
% ACL 2012 by Maggie Li and Michael White,
% ACL 2010 by Jing-Shin Chang and Philipp Koehn,
% ACL 2008 by Johanna D. Moore, Simone Teufel, James Allan, and Sadaoki Furui,
% ACL 2005 by Hwee Tou Ng and Kemal Oflazer,
% ACL 2002 by Eugene Charniak and Dekang Lin,
% and earlier ACL and EACL formats written by several people, including
% John Chen, Henry S. Thompson and Donald Walker.
% Additional elements were taken from the formatting instructions of the \emph{International Joint Conference on Artificial Intelligence} and the \emph{Conference on Computer Vision and Pattern Recognition}.

\bibliography{ACL}

\appendix

\section{Appendix}
\label{sec:appendix}

\begin{table*}[t]
    
  \setlength{\tabcolsep}{4.2 pt}
  \small
  \centering

  \begin{tabular}{lcccccccccccccc}

  \toprule
  
  \bf Model & \bf dataset & \bf epoch & \bf optimizer & \bf learning scheduler  & \bf learning rate & \bf warm up steps & \bf LoRA rank     \\
  \midrule

  Llama-3b~(SFT) & LongOrca  & 1 & AdamW & linear & 2e-5 & 0& 32  \\
   
  \midrule
  
  DeciMamba~(SFT) & LongOrca & 1 & AdamW & linear & 2e-5 & 0& 32\\ 
  Mamba~(SFT) & LongOrca & 1 & AdamW & linear & 2e-5 & 0& 32\\ 
  ReMamba~(SFT)  & LongOrca & 1 & AdamW & cosine & 2e-5 & 0& 32 \\
  Mamba2~(SFT) & LongOrca & 1 & AdamW & cosine & 2e-5 & 0& 32\\ 
  ReMamba2~(SFT)  & LongOrca & 1 & AdamW & cosine & 2e-5 & 0& 32 \\

  \bottomrule
  \end{tabular}
  
  \caption{Model training details
  % Llama-2-3B's finetuning process uses the same dataset as Mamba and ReMamba but truncates input sequences to 2K tokens during training, which proves to be the best configuration.
  }
  \label{tab: detail}
\end{table*}

\begin{table}[t]
    
  \setlength{\tabcolsep}{3.5pt}
  \small
  \centering
  \begin{tabular}{lcccccccccccccc}
  \toprule
  \bf Model & \bf $\rho$ & \bf $p$   & \bf $s$ & \bf model\_type & \bf average  \\
  \midrule
  ReMamba  &0.009 & 0.18 & 0.00 & ReMamba  & 27.86 \\
  middle0.0 &0.009 & 0.18 & 0.00 & middle  & 25.96 \\
  middle0.1 &0.009 & 0.18 & 0.10 & middle  & 26.45 \\
  middle0.2 &0.009 & 0.18 & 0.20 & middle  & 26.86 \\
  middle0.3 &0.009 & 0.18 & 0.30 & middle  & 26.56 \\
  middle0.4 &0.009 & 0.18 & 0.40 & middle  & 26.43 \\
  special   &0.009 & 1.00 & 1.00   & special  & 15.76 \\
  \bottomrule
  \end{tabular}
  
\caption{Performance of different model variants on LongBench. Parameters $s$, $p$, and $\rho$ represent the relative start position, relative length to compress, and compression ratio, respectively. In this context, the ``ReMamba'' model type constitutes our optimal model. The ``middle'' type corresponds to the model variant where $s$ is non-zero. The ``special'' model variant compresses the entire prompt using $\rho=0.009$ and subsequently appends the compressed hidden states to the end of the original prompt in the second stage.}
\label{tab:middlelongbench}
\end{table}

\subsection{Training Details}
\label{sec:appendixdetail}
During training, the ReMamba hyperparameter  $s$ is fixed at 0. The hyperparameter $p$ is randomly sampled from the interval [0.1, 0.3], while $\rho$ is randomly sampled from the interval [0.05, 0.2].

Most of DeciMamba's experiments are conducted on Mamba130m. The only configuration for DeciMamba2.8 provided in the original paper is decimating\_layers = 22, decimation\_max\_p\_L\_base = 4000, used for language modelling tasks. Here we change the decimation\_max\_p\_L\_base = 6000. For all other hyperparameters, we use the default settings: decimation\_min\_seq\_len = 20, decimation\_beta = 0.5, and decimation\_type = ``max\_p``.
The linear positional interpolation configurations for Llama-3B are: max\_position\_embeddings = 6144 and factor = 3.
The data construction process for Llama-3b is identical to that of Mamba.
We finetune all the models for 1 epoch using DeepSpeed Zero Stage 3 on 8 A100-80GB GPUs. Other training details can be found in the Table \ref{tab: detail}.

\subsection{Mamba2 Details}
\label{sec:appendixmamba2detail}
The hyperparameters here are: $s=0$, $p=0.25$ and $\rho=0.05$. The max length is still 6k. It is noteworthy that Mamba2 exhibits nearly no performance improvement over Mamba on LongBench, suggesting potential limitations within the Mamba model series.

\subsection{Why compress from the start}
\label{sec:appendixs}
Experimental results indicate that setting $s=0$ is the best. 
However, one might wonder about the effectiveness of compressing in the middle of the sequence. We conduct additional analytical studies to explore the impact of compressing the input sequence from different starting positions.

We train a model utilizing $s$ sampled uniformly from the interval [0.1, 0.3] during the training process. Subsequently, we evaluate its performance on LongBench under conditions identical to those of the ReMamba model, employing a maximum length of 6k tokens, $p=0.18$, and $\rho=0.009$. We evaluate the average scores ranging $s$ from 0 to 0.4. Additionally, we train a special model variant that compresses the entire prompt based on $\rho = 0.009$ and appends the compressed hidden states to the end of the original prompt in the second stage.

% As is shown in Table \ref{tab:middlelongbench}. We can \bei{see that} a degradation in performance if we compress in the middle. The extreme branch is even worse than the sft Mamba baseline.
Table \ref{tab:middlelongbench} presents the results of these experiments. We observe a performance degradation when the compression is applied in the middle of the sequence. The special model variant performs even worse than the finetuned Mamba baseline.
 
This degradation can be explained by the disruption caused to the causal language modeling nature of the Mamba model. When compressed information is integrated into the initial position, the subsequent language modeling process can proceed without modification, effectively treating the compressed data as a specialized non-zero initial state. Conversely, inserting those compressed hidden states as tokens within the sequence disrupts the causal language modeling paradigm, which assumes complete sentences as input. This incongruity hinders the model's ability to maintain a coherent state space and can lead to performance degradation. Among the tested models, the special model variant that appends compressed hidden states to the end of the original prompt exhibits the most pronounced negative impact due to the significant disruption of the model's expected input structure.

Despite these challenges, the model that compresses in the middle still outperforms the finetuned Mamba baseline. This demonstrates that our method exhibits apparent effectiveness.

\end{document}